\documentclass[letterpaper, 10 pt, conference]{ieeeconf}  

\IEEEoverridecommandlockouts                              

\usepackage{diagbox}
\usepackage{color}
\usepackage{gensymb}
\usepackage{times}
\usepackage{epsfig}
\usepackage{tabularx}
\usepackage{amssymb}
\usepackage{bbm}
\usepackage[misc]{ifsym}
\usepackage{svg-extract}
\let\labelindent\relax
\usepackage{enumitem}
\usepackage{calc}
\usepackage{microtype}
\usepackage{amsmath}
\usepackage{graphicx}
\usepackage{subcaption}
\usepackage{multirow}
\usepackage{booktabs,colortbl,hhline,mathtools,makecell}
\usepackage{bm}

\usepackage{graphicx}
\usepackage{rotating}
\usepackage{svg}
\usepackage{microtype}
\usepackage{yhmath}

\graphicspath{{figures/}}
\svgpath{{figures/}}

\usepackage[colorlinks]{hyperref}  

\title{\bf
FRESHR-GSI: A Generalized Safety Model and Evaluation Framework for Mobile Robots in Multi-Human Environments}

\author{Pranav Pandey \and Ramviyas Parasuraman \and Prashant Doshi
\thanks{The authors are with the School of Computing, University of Georgia, Athens, GA 30602, USA. 
Author emails: {\small \{pranav.pandey,ramviyas,pdoshi\}@uga.edu}}}

\DeclareMathOperator*{\argmax}{arg\,max}

\begin{document}

\maketitle

\begin{abstract}
Human safety is critical in applications involving close human-robot interactions (HRI) and is a key aspect of physical compatibility between humans and robots. While measures of human safety in HRI exist, these mainly target industrial settings involving robotic manipulators.  
Less attention has been paid to settings where mobile robots and humans share the space. This paper introduces a new robot-centered directional framework of human safety. It is particularly useful for evaluating mobile robots as they operate in environments populated by multiple humans. The framework integrates several key metrics, such as each human's relative distance, speed, and orientation. The core novelty lies in the framework's flexibility to accommodate different application requirements while allowing for both the robot-centered and external observer points of view. We instantiate the framework by using RGB-D based vision integrated with a deep learning-based human detection pipeline to yield a generalized safety index (GSI) that instantaneously assesses human safety. We evaluate GSI's capability of producing appropriate, robust, and fine-grained safety measures in real-world experimental scenarios and compare its performance with extant safety models. 
\end{abstract}

\section{Introduction}
\label{Sec:introduction}

The study of human-robot interaction (HRI) and collaboration has gained importance as more humans and robots share workspaces and engage in proximal encounters~\cite{lasota2017survey,veloso2018increasingly}. Robot co-workers can significantly enhance productivity and efficiency in repeatable and collaborative tasks. For this, the interacting human and robot must be physically compatible, and a key aspect of such compatibility is human safety~\cite{khalid2017safety}.

A substantial body of literature focuses on human safety in industrial settings \cite{valori2021validating,matheson2019human}, with established safety criteria and guidelines for collaborative robots \cite{harper2010towards,palmieri2024control,nertinger2023influence} including industrial robot safety standards such as ISO 10218 and ISO/TS 15066 \cite{valori2021validating,ferraguti2020control}. Researchers have also introduced real-time safety assessments for large manipulators in human-robot collaborations~\cite{kulic2007pre,lippi2018safety,palmieri2024control}.
While appropriate for close proximity and contact interactions, such as in manufacturing, these safety standards and measures
cannot be readily transferred to mobile robots, which typically operate in large, unbounded workspaces, where the mobility of humans and robots significantly impacts human safety. Furthermore, existing methods in the literature tend to underestimate safety in multi-human environments. 
As such, there is a need for robust measures of safety that can be obtained from different points of view (proprioceptive/exteroceptive) and for different utilities, such as safety assessments and motion control.

\begin{figure}[t]
    \centering
    \includesvg[width=0.98\linewidth]{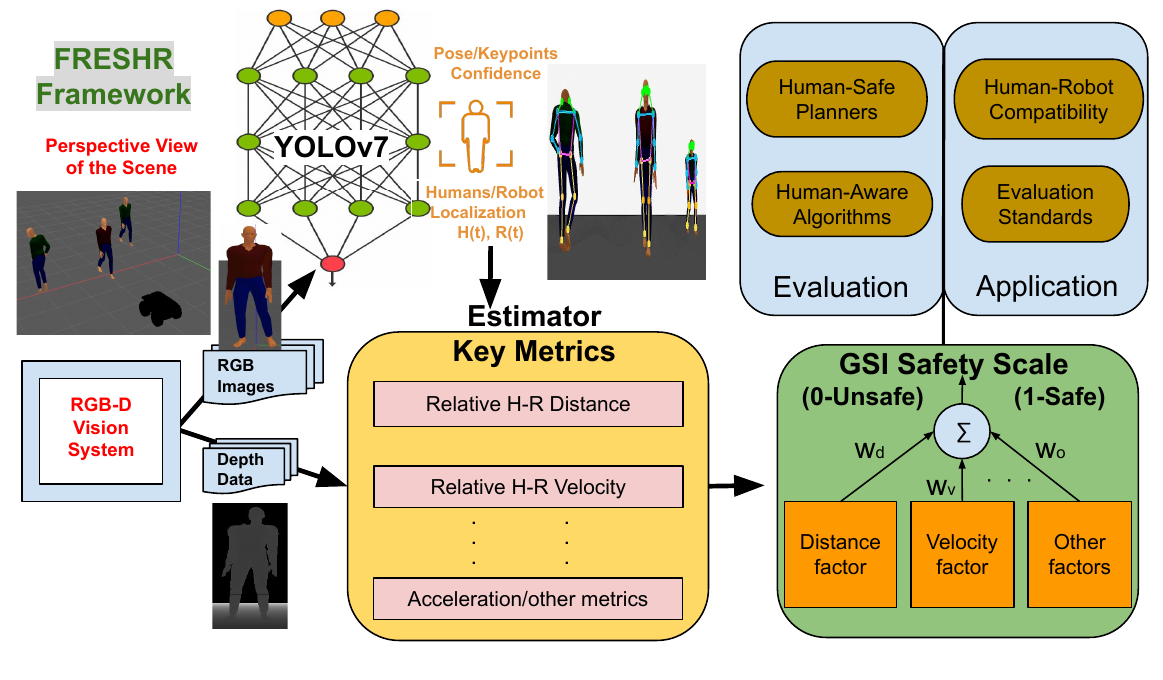}
    \vspace{-3mm}
    \caption{\small Overview of the proposed vision-based safety evaluation framework (FRESHR). A continuous sequence of RGB images is supplied to Yolov7 for human and robot detection and human pose estimation. These detections and their confidence values will be integrated with depth information to calculate key metrics such as the relative distance and velocity between humans and robots. Finally, a normalized safety value will be provided to evaluate the multi-human safety of mobile robots.}
    \label{fig: framework}
    \vspace{-3mm}
\end{figure}

In this paper, we introduce a FRESHR framework\footnote{Available: \url{https://github.com/herolab-uga/FRESHR-GSI}} for evaluating the physical safety of humans in shared human-robot workspaces, labeled as FRESHR. The framework integrates multi-modal data collected using depth cameras mounted either on the mobile robot or on an external observer of the HRI to yield a human safety value called the generalized safety index (GSI). We use a directional model designed for mobile robot applications that combines the impact of distance, velocity, acceleration, and the angular range between the robot and nearby humans. GSI is particularly useful in an environment populated by multiple humans, prioritizing the safety of those at greater risk.  

FRESHR sets itself apart from other vision-based measurement systems by enabling safety evaluation from multiple points of view: from the robot -- which enables human-aware control, from an external observer of the HRI or another robot in a multi-robot system -- to assess the safety of humans interacting with the other robots. GSI is a robust and fine-grained safety assessment, differing from extant scales, 
which mainly focus on close proximity collaborations between a {\em single} human and a manipulator. In contrast, FRESHR is amenable to being used in crowded environments. 
Demonstration of FRESHR in real-world environments is shown in the attached video\footnote{\url{https://youtu.be/c8eoYeW05UE}}.

These contributions are pivotal to motivating and stimulating further innovations in improving human safety for mobile robots and achieving human-safe robot control.

\section{Related Work}
\label{Sec: Related}

 Interpersonal distance studies have resulted in the development of various scientific terms across different disciplines. One of the most recognized terms is "proxemics," introduced by Hall \cite{hall1966hidden}. Proxemics broadly investigates spatial relationships between individuals and provides a framework for understanding how people use and manipulate space according to their preferred levels of closeness and interaction. Hall's theory identifies four primary distance zones: the intimate zone (0–0.46m), the personal zone (0.46–1.2m), the social zone (1.2–3.7m), and the public zone ($>$3.7m), which have been widely applied in many Human-Robot Interaction studies \cite{kamide2014direct, rossi2017user, yasumoto2011personal, brandl2016human}.
Numerous findings involving human subject experiments and surveys \cite{leichtmann2020much, kamide2014direct, yasumoto2011personal} have corroborated that humans perceive an interaction with an approaching robot as safe if it can stop in the personal zone and unsafe if it is (or about) to breach the intimate zone. Motivated by these findings, an appropriate safety measure should provide a granular value of safety level based on whether the robot can stop within the public zone (safe) and never breach any human's intimate zone (unsafe).

In the literature, several models measure safety and performance in human-robot interaction \cite{steinfeld2006common}, but few address the safety features of mobile robots, particularly in dynamic, human-shared workspaces \cite{truong2017toward}. Unlike industrial solutions restricted to safety zone monitoring, vision-based speed, and separation monitoring systems offer ubiquity, transferability, and ease of deployment \cite{halme2018review}. 
For instance, \cite{rodrigues2022modeling} used a vision-based system with deep learning models for collision detection and decision support in safe human-robot collaboration. 
Maria et al. \cite{maria2022vision} used a similar approach, detecting and tracking humans in industrial HRI with zones based on human speed and robot reaction time. RGB-D cameras enhance detection capabilities with depth data. In \cite{svarny2019safe}, keypoint data were used to implement speed and separation monitoring alongside power and force limiting. Tashtoush et al. \cite{tashtoush2021human} utilized a top-view (exteroceptive) RGB-D camera for precise operator positioning near a robotic manipulator, and Secil and Ozkan \cite{secil2022minimum} proposed skeletal tracking of human motion and obtaining the minimum distance as an alternative to wearable systems in safety measures. 
However, these methods are often limited to static exteroceptive perspectives of safety measures rather than being applied to mobile robots.

Innovative approaches for ensuring human safety in human-robot co-shared workspaces include planning and control strategies for manipulators using human monitoring data.
Traditional single-human interaction models fall short of addressing the complexities of real-world applications such as service delivery, search and rescue, logistics, and warehousing, where multiple humans are present in the robot's operational domain. Existing methods often overlook critical factors like velocity, orientation, and the feasibility of operation in open environments with coexisting mobile robots and humans.
Kulic and Croft \cite{kulic2007pre} defined a Danger Index (DI) based on a product formulation of human-robot distance and velocity for safe trajectory planning in robot arms. Lacevic et al. \cite{lacevic2013safety} developed the Kinetostatic Danger Field (KDF) for real-time danger assessment and control adjustments. Lippi et al. \cite{lippi2018safety} extended KDF to a human-safety assessment (HSA) for multi-robot collaboration, adjusting paths based on human proximity. Palmieri et al. \cite{palmieri2024control} presented a human safety field (HSF) control architecture for enhancing safety in shared workspaces by adjusting manipulator trajectories.
However, these models are intended for industrial manipulators and often fall short in dynamic, unbounded spaces. As we show later in Sec.~\ref{sec:experiments}, these methods can result in misleading safety assessments, either underestimating risk or falsely indicating safer conditions due to their reliance on the summative or product-based integration of different metrics like distance and velocity.  
These issues are more pronounced when estimating safety in multi-human environments, where each human do not contribute equally to the safety assessment.
For example, in the product-based integration of distance and velocity metrics for risk assessment \cite{kulic2007pre}, if the robot is in the human's intimate zone but not moving, the DI will assess this situation as non-risky. 
Similarly, summation-based scales where the safety level of multiple humans around a robot is estimated through an averaging approach \cite{palmieri2024control} can potentially overestimate the safety level due to the influence of humans in safe zones dominating the critical risk posed to humans in unsafe zones. These limit their applicability in assessing and ensuring human safety around robots. 
In light of these findings, it is evident that despite significant advancements, assessing and ensuring safety for mobile robots in dynamic, shared human environments remains challenging. 

\section{FRESHR: Evaluating Human Safety}
\label{sec:GSI}

Let a mobile robot $r$ with pose, $p_r = \langle \bm{x}_r,\theta_r\rangle$ where $\bm{x}_r = (x_r, y_r, z_r)$, be co-located with multiple humans, $\{ h_i | i = 1 \ldots N_h\}$ where $N_h$ is the number of detected humans. Let a human $h_i$ be detected at a position $\bm{x}_{h_i} = (x_i,y_i,z_i)$ in a common frame of reference. The frame may be centered on the robot or global based on an external observer. The problem facing FRESHR is determining the current level of human safety as influenced by the robot $r$. 
We assume that the robot's motion constraints, such as its maximum speed in any direction $V_{max}$, and maximum deceleration $A_{max}$, are known and that the robot has sensors to detect and localize multiple humans within a limited sensor range and field of view and able to estimate the relative distance and velocities between every detected human and the robot (see Sec.~\ref{sec:framework}). We must assess the physical safety of the humans around the robot in its direction of travel $\theta_r$.

\subsection{Generalized Safety Index}
\label{sec:HSI}

In our framework, three key components are integrated to assess the safety of every detected human: distance, relative velocity, and the bearing of the human from the robot. These measures are generally deemed sufficient for assessing safety within the interaction space~\cite{nertinger2023influence}. 

For each human $h_i$ at position $\bm{x}_{h_i}$ in a common reference frame, let $d_{h_i,r}$ denote the distance from the robot, $d_{h_i,r} = \|\bm{x}_{h_i} - \bm{x}_r\|_2$. The relative velocity between them is the first-order derivative of the distance, $v_{h_i,r} = -\dot{d}_{h_i,r}$, which is a positive value when the human and robot move towards each other and a negative value otherwise. Denote the relative bearing of the human $h_i$ w.r.t. the robot as $\theta_{h_i,r} = \measuredangle(\bm{x}_{h_i} -\bm{x}_{r}) - \theta_r $. To clarify, this bearing is the angle (measured counterclockwise from the positive x-axis) between the segment joining the robot to the human and the robot's current orientation $\theta_r$. 

\begin{figure}[!ht]
    \centering
    \vspace{-2mm}
    \includegraphics[width=0.90\linewidth]{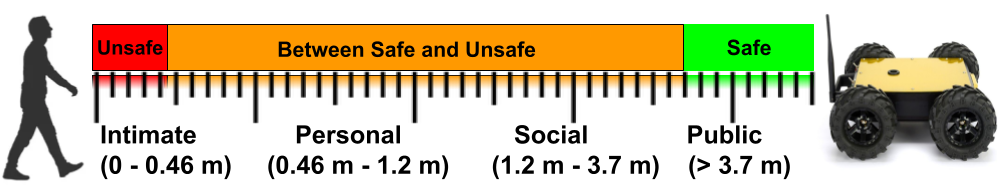}
    \caption{\small FRESHR aligns its safety scale with the known and empirically determined proximity ranges in human-robot interaction spaces. GSI takes a value of 1 indicating safe (green) in the public space, (0 - 1] (amber) in the personal and social spaces, and 0 (red) in the intimate space.}
    \label{fig:space_diag}
\end{figure} 

To arrive at an appropriate safety scale, we rely on the well-known concepts of intimate, personal, social, and public spaces of human-robot interactions~\cite{hall1966hidden}, which are illustrated in Fig.~\ref{fig:space_diag}. Generally, a human's intimate space was empirically determined to be a sphere of radius 0.46m centered on the human, her personal and social spaces are the spherical shells whose radius lies in (0.46m - 1.2m] and (1.2-3.7m] ranges respectively, and the region beyond 3.7m is considered a public space. Our approach is to assess human safety based on where and whether the mobile robot intrudes into these spaces (i.e., the stopping zone). 
Towards this, we define a generalized safety index of a human $h_i$ as 
\begin{small}
\begin{equation}
    \widehat{GSI}_{h_i}(d_{h_i,r},v_{h_i,r}; \rho) = 
        \left[ \frac{d_{h_i,r} - \left(\text{s}(v_{h_i,r}) \frac{v_{h_i,r}^2}{2 A_{max}} + D_{min} \right)}{D_{max} - D_{min}}\right]^{\rho}
    \label{eqn:GSI_hat}
\end{equation}  
\end{small}
Here, $A_{max}$ is a constant representing the maximum (de-)acceleration of the robotic platform; 
$D_{max}$ is the distance beyond which the human's safety is assured -- we may let $D_{max} = 3.7m$ (public space); and a mobile robot should not come closer than $D_{min}$ -- we may let $D_{min}=0.46m$ (intimate space), or 0 if, for example, the robot needs to transport the human. The term $\frac{v_{h_i,r}^2}{2 A_{max}}$ in \eqref{eqn:GSI_hat} indicates the distance required for the robot to come to a stop from its current relative speed $v_{h_i,r}$, given a maximum deceleration rate of $A_{max}$. $\text{s}(v_{h_i,r})$ is the sign function informing whether the human is approaching or moving away from the robot.

\begin{figure}[!ht]
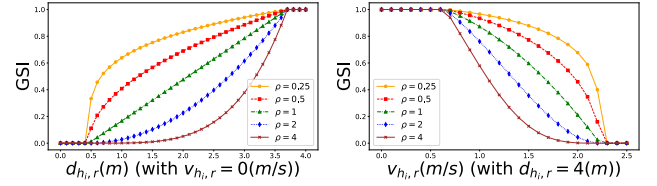
 
    \centering
    \vspace{-2mm}
    \includesvg[width=0.48\linewidth]{rho_graph_dist.svg}
    \includesvg[width=0.48\linewidth]{rho_graph_vel.svg}
    \vspace{-2mm}
    \caption{\small GSI can be fitted to various applications, robot platform properties, and subjective safety perceptions of humans through parameter $\rho > 0$. For instance, $\rho=1$ is set for assessing safety, $\rho>1$ for more cautious robot control, and $\rho<1$ for a more closer interaction with human who are already comfortable.}
    \label{fig: variation_graph}
    \vspace{-3mm}
\end{figure}

The hyperparameter $\rho > 0$ provides a way to fit GSI to various kernels based on the current application setting and the subjective human perception of safety. We may select different values of $\rho$ in applications involving GSI-aided motion control, where higher $\rho>1$ decay of safety can be appropriate in robots with slow reaction times or large mass (i.e., a larger than usual buffer from the human is preferred for more cautious human perceptions of safety or higher chances of a platform failure to stop in fast motion settings~\cite{lasota2014toward}. 
Previous work has utilized a similar parameter for industrial robots, where it is set to 2~\cite{kulic2007pre}. On the other hand, lower values of $\rho<1$ may be utilized if the human is comfortable around mobile robots \cite{edelmann2023interaction}, reducing the need for unnecessary interventions~\cite{kulic2007pre}. Finally, a balanced trend can be obtained with $\rho=1$ providing a rational GSI \cite{lasota2014toward}, and therefore, we use this setting ($\rho=1$) for assessing the current safety level. Fig.~\ref{fig: variation_graph} illustrates the impact of $\rho$ on the GSI model.

In essence, $\widehat{GSI}_{h_i}$ accounts for the robot's ability to stop before breaching the intimate zone of a human. A value $\leq 0$ represents an unsafe condition (i.e., the intimate space has or is about to be breached), while GSI $\geq 1$ asserts a fully safe condition (i.e., the robot is in the public zone). Any value between 0 and 1 measures the safety level, closer to 0 indicates less safety, and higher risk to the human at that point in time, whereas closer to 1 suggests that the human is likely to be safe at that time. 

\begin{figure}[!ht]
    \centering
    \begin{minipage}{3.3in}
    \includegraphics[width=0.4\linewidth]{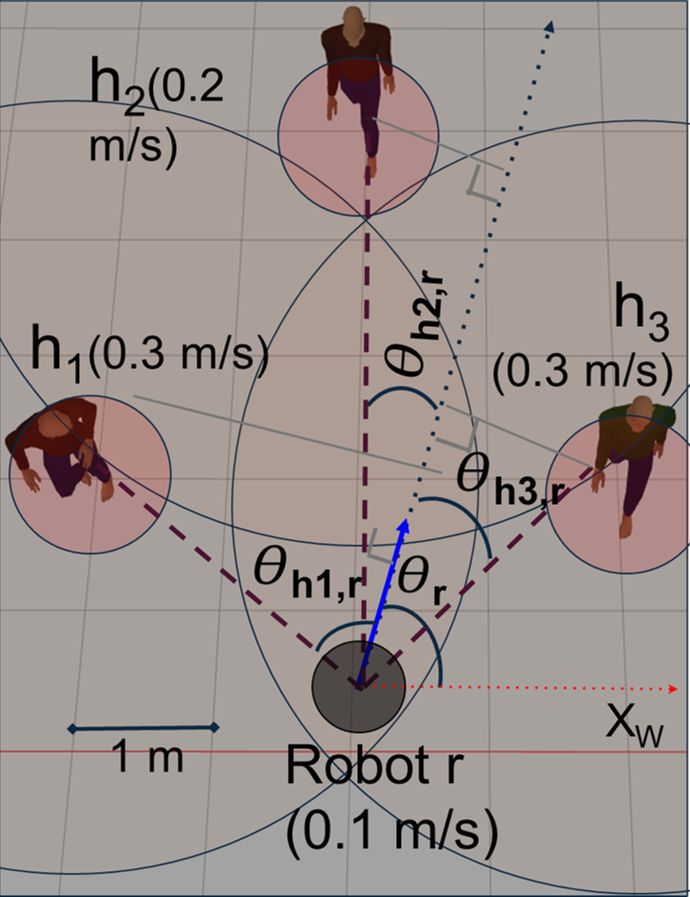} 
    \includesvg[width=0.59\linewidth]{figures/circular_graph.svg}
    \centerline{\small ($a$)}
    \end{minipage}
    \begin{minipage}{3.3in}
    \centerline{\includesvg[width=0.75\linewidth]{figures/tau_values.svg}} 
    \vspace{-5mm}
    \centerline{\small ($b$)}
    \end{minipage}
    \caption{\small ($a$) An example setting with three humans in the vicinity of the mobile robot $r$. FRESHR yields a directional safety value for each human. In this example, $GSI_{h_1} = 0.7$ with $\theta_{h_1,r} = 290^\circ $, $GSI_{h_2} = 0.9$ with $\theta_{h_2,r} = 345^\circ $, and $GSI_{h_3} = 0.4$ with $\theta_{h_3,r} = 30^\circ $, each of which is calculated using Eq.~\ref{eqn:gsi_single_thetar}. ($b$) Impact of the hyperparameter $\tau$ on the collective $GSI$ (polar plot in (a)) obtained using Eq.~\ref{eqn:gsi_overall}. The individual GSIs used are those from ($a$).}
    \label{fig:directionalGSI}
    \vspace{-6mm}
\end{figure}

Eq.~\ref{eqn:GSI_hat} is applicable for a static scenario or if we assume the robot and human are directly approaching each other, i.e. $\theta_{h_i,r} = 0^\circ$. For a non-zero bearing of the human w.r.t. the robot, we extend Eq.~\ref{eqn:GSI_hat} to scale the GSI with how close the robot gets to the human as it passes by it. More specifically, we obtain a directional GSI in FRESHR as given below,      
\begin{equation}
GSI_{h_i} (d_{h_i,r},v_{h_i,r},\theta_{h_i,r};\rho) = 1 - (1 - \widehat{GSI}_{h_i})\cos\theta_{h_i,r}.
\label{eqn:gsi_single_thetar}
\end{equation}
We illustrate the derivation of Eq.~\ref{eqn:gsi_single_thetar} using  Fig.~\ref{fig:directionalGSI}($a$), which shows that $\cos\theta_{h_i,r}$ can be used to scale the complement of the GSI value that is obtained as if the robot is heading straight for the human. Notice that when $\theta_{h_i,r} = 0$, $\cos\theta_{h_i,r} = 1$ and $GSI_{h_i}(d_{h_i,r},v_{h_i,r},\theta_{h_i,r};\rho)$ collapses to $\widehat{GSI}_{h_i}(d_{h_i,r},v_{h_i,r};\rho)$ as we may expect. And, if $\widehat{GSI}_{h_i}(\cdot)$ indicates not safe, then $GSI_h(\cdot)$ tempers down the non-safety by how close the robot is expected to pass by the human. Thus, GSI represents a dynamic measure of safety, integrating real-time motion input to assess the human's safety in the shared workspace given the robot's movement.

\vspace{0.1in}
\noindent \textbf{GSI for settings shared with multiple humans}
Implications of robot motion on the safety of multiple humans (e.g., in crowded pedestrian areas~\cite{salvini2022safety}) are studied from motion planning and physiological social awareness perspectives~\cite{truong2017toward,ferrer2013robot}. The presence of multiple humans in the shared workspace complicates the determination of safety as we now face an additional challenge: how to aggregate individual safety indications to determine the safety of the whole.


Previous work~\cite{palmieri2024control} has averaged the individual safety values over all humans. This approach has the disadvantage that it may {\em overestimate} the overall safety when the robot is safe for a majority of the humans in the group but unsafe for a few in the shared space. Therefore, we posit that the safety index for the whole should not only be directional but should also attribute higher importance to the safety of those humans for whom the robot presents significant safety implications in its intended direction. Toward this, let ${\bm d}_{h,r} = \langle d_{h_i,r} | i = 1, \ldots, N_h \rangle$ represent the vector of relative distances between the mobile robot and each human $i$ in the shared space and analogously
${\bm v}_{h,r}$ and ${\bm \theta}_{h,r}$ represent the vector of relative velocities and angles, respectively. 
%
Rather than simply returning the minimum of the $GSI_h(\cdot)$ values, we utilize a smooth minimum LogSumExp (also known as the realsoftmin) of the individual values, to obtain the {\em collective GSI} for the group of $N_h$ humans in the robot's shared space. 
\begin{equation}
\begin{split}
& GSI({\bm d}_{h,r},{\bm v}_{h,r};{\bm \theta}_{h,r}; \rho, \tau)   = \\ 
&   -\tau \ln \left ( \frac{1}{N_h} \sum\limits_{i=1}^{N_h} e^{\frac{-GSI_{h_i}(d_{h_i,r},v_{h_i,r},\theta_{h_i,r};\rho)}{\tau}} \right)  
\label{eqn:gsi_overall}
\end{split}
\end{equation}
where $GSI_h(\cdot)$ is as defined previously in Eq.~\ref{eqn:gsi_single_thetar} for a single human in the vicinity, $\tau$ is a hyperparameter that controls the smoothness of the approximation of the minimum of the $GSI_h(\cdot)$ values. As $\tau$ reduces, $GSI$ converges to the minimum $GSI_h(\cdot)$ across all humans $i$. We set $\tau=0.01$ for obtaining close to the absolute minimum. The LogSumExp function heavily penalizes larger $GSI_h(\cdot)$, which makes it sensitive to the small $GSI_h(\cdot)$ values, thereby obtaining a safety index corresponding to the human that influences the most. 
GSI in Eq.~\ref{eqn:gsi_overall} satisfies the core properties of safety measures~\cite{lippi2018safety,lacevic2013safety} such as a monotonic increase (decrease) with distance (velocity) and differentiability. These properties enable a differential safety scale that is useful in evaluating and integrating mobile robot algorithms.
For instance, similar to \cite{lacevic2013safety}, GSI can be vectorized for control applications as
\begin{equation}
   \overrightarrow{GSI} (t) = {GSI}({\bm d}_{h,r},{\bm v}_{h,r};{\bm \theta}_{h,r},t; \rho) . \frac{\nabla {GSI}
(t)}{\|\nabla {GSI}(t)\|} ,
    \label{eqn:gsi_vector}
\end{equation}
where $\nabla$ is either the time-based or pose-based derivative. Finally, if the collective also includes multiple mobile robots, FRESHR can further generalize to obtain the GSI that is a smooth minimum across the collective GSIs w.r.t. all the robots.  We summarize GSI for various scenarios in Table~\ref{tab:scenarios_appropriateness} and compared with other scales for appropriateness.

\begin{table*}[!t]
\caption{\small A summarization of the appropriateness of different safety scales in various scenarios (combinations of distance $d_{h_i,r}$ and relative velocity $v_{h_i,r}$). The appropriate safety level is determined based on the stopping zone of the robot to the closest human (Fig.~\ref{fig:space_diag}). SH - single human. MH - multi-human. A \checkmark or $\times$ indicates whether the scale correctly informs the safety level. }
\label{tab:scenarios_appropriateness}
\centering
\resizebox{\linewidth}{!}{
\begin{tabular}{|c|c|c|c|c|c|c|c|c|c|c|c|c|c|c|}
\hline
\textbf{Scenario} & \textbf{Distance} & \textbf{Relative Velocity} & \textbf{Stopping Zone} & \textbf{Appropriate Assessment} & \multicolumn{2}{c|}{\textbf{GSI} [Ours]} & \multicolumn{2}{c|}{\textbf{DI} \cite{kulic2006real}} & \multicolumn{2}{c|}{\textbf{KDF} \cite{lacevic2013safety}} & \multicolumn{2}{c|}{\textbf{HSF} \cite{palmieri2024control}}  & \multicolumn{2}{c|}{\textbf{HSA} \cite{lippi2018safety}}  \\
\hline
& & & & & SH & MH & SH & MH & SH & MH & SH & MH & SH & MH \\
\cline{6-15}
A & $d_{h_i,r} \geq D_{max}$ & $v_{h_i,r} \leq 0 $ & Public & Safe & \checkmark & \checkmark & \checkmark & N/A & \checkmark & \checkmark & \checkmark & \checkmark & \checkmark & \checkmark \\
B & $d_{h_i,r} \geq (D_{max} + \frac{v_{h_i,r}^2}{2 A_{max}})$ & $v_{h_i,r} \geq 0$ & Public & Safe & \checkmark & \checkmark & \checkmark & N/A & \checkmark & \checkmark & \checkmark & \checkmark & \checkmark & \checkmark \\
C & $d_{h_i,r} \geq D_{max}$ & $0 < v_{h_i,r}^2 < 2A_{max}(d_{h_i,r} - D_{min})$ & Within Personal/Social & Between & \checkmark & \checkmark  & $\times$ & N/A & \checkmark & \checkmark & $\times$ & $\times$ & \checkmark & \checkmark \\
D & $D_{min} \leq d_{h_i,r} \leq D_{max} $ &  $v_{h_i,r} = 0 $ & Within Personal/Social & Between & \checkmark  & \checkmark  & $\times$ & N/A & \checkmark & $\times$ & \checkmark & \checkmark & \checkmark & \checkmark\\ 
E & $d_{h_i,r} \geq D_{max}$ & $v_{h_i,r}^2 \geq 2A_{max}(d_{h_i,r} - D_{min})$ & Intimate & Unsafe & \checkmark  & \checkmark  & $\times$ & N/A & \checkmark & $\times$ & $\times$ & $\times$ & $\times$ & $\times$ \\
F & $d_{h_i,r} \leq (D_{min} + \frac{v_{h_i,r}^2}{2 A_{max}})$ & $v_{h_i,r} \geq 0$ & Intimate & Unsafe & \checkmark  & \checkmark  & $\times$ & N/A & \checkmark & $\times$ & $\times$ & $\times$ & \checkmark & \checkmark\\
\hline
\end{tabular}
}
\end{table*}

\subsection{FRESHR Implementation}
\label{sec:framework}

We develop a real-time RGB-D-based system to measure robot safety during interactive tasks and integrate this with our GSI scale (see Fig.~\ref{fig: framework}). The use of depth point clouds for obtaining the relative distance between the robot and humans enable deployment flexibility from different viewpoints, such as external agents or onboard robot sensors, ensuring safety and comfort for humans. 
In it core, the framework employs a deep learning pipeline for detecting humans and robots and uses algebraic calculations to extract safety-related factors. We utilize YOLOv7 \cite{wang2022yolov7} for real-time object detection and localization from RGB images, which also provides skeleton keypoint locations for whole-body safety. These detections are correlated with depth values from synchronized images.

YOLOv7 provides confidence scores for each detection, useful for integrating multiple skeletal keypoint distances. Detected pixel locations are converted to world frame coordinates using the OpenCV library, applying the camera's intrinsic matrix. This is integrated with depth information to estimate real-world 3D coordinates for each object (human position $p_h$, robot position $p_r$, and relative orientation $\theta_{h_i,r}$). The detected skeletal keypoints are combined by weighted addition $d_{hr} = \sum_k w_k d_k$, with confidence scores as weights $w_k$ for the $k^{th}$ keypoint at a relative distance of $d_k = \|p_r - k\|$. Scores are normalized such that $\sum_k w_k = 1$. Keypoints passing a confidence threshold $conf_{thr}$ allow us to remove noisy detections. A confidence threshold 0.9 is used to obtain the best detection accuracy.

The camera frame acts as the global reference frame, and ${\bm d}_{h,r},{\bm v}_{h,r};{\bm \theta}_{h,r}$ in the robot's frame can be obtained using the camera's rotation and translation matrices if the camera is rigidly mounted on the robot. Euclidean norm is used for distances.
For geometric rigidity, we use three distance measurements for robust velocity estimates, i.e., ${v}_{hr} = \sqrt{\frac{d^2_{hr}(t-2) - 2d^2_{hr}(t-1) + d_{hr}^2(t)}{2}}/(T)$, where $\frac{1}{T}$ is the measurement frequency. If no detections are made (e.g., low detection confidences, out of sensor range, or due to occlusions), the estimates will not be used at that time.

\section{Experimental Evaluation}
\label{sec:setup}

We conduct extensive experiments to validate different aspects of FRESHR. First, we analyze the framework's capability to provide acceptable safety assessments and compare GSI with extant scales in the literature. 
We present an experiment in a multi-human setting with a physical mobile robot running  FRESHR using its onboard sensors and computer. Next, we obtain GSI in simulations and real-world crowd robot datasets, illustrating GSI's potential use in evaluating human-aware motion planners.

\begin{figure}[!t]
    \begin{minipage}{3.4in}
    \centering
    \includesvg[width=0.48\linewidth]{odom1.svg} 
    \includegraphics[width=0.48\linewidth]{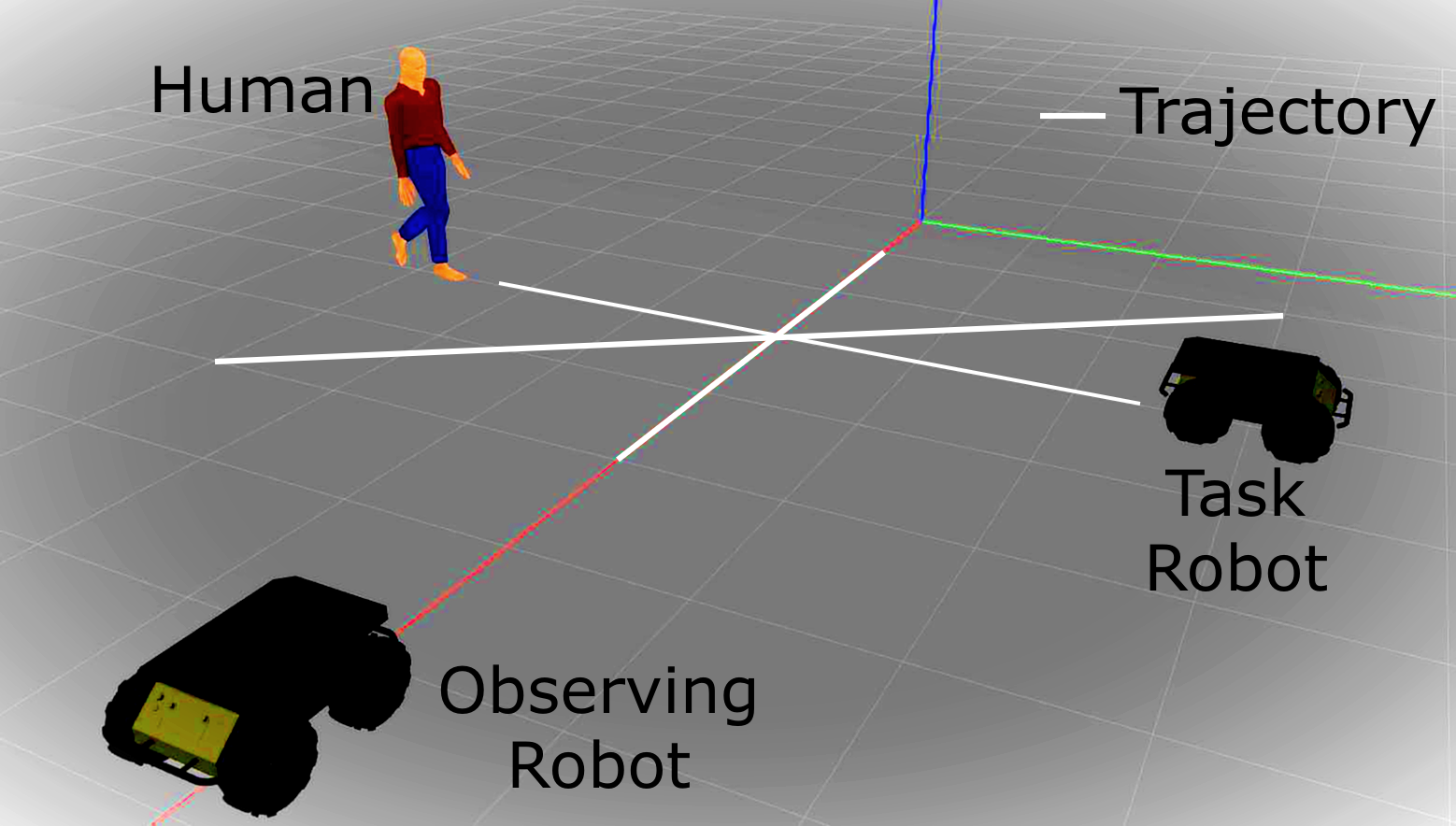} 
    \centerline{\small ($a$)}
    \end{minipage}
    \begin{tabular}{cc}
    \multicolumn{2}{c}{\textbf{\small Task robot's viewpoint}} \vspace{-1mm}\\
    \includegraphics[width=0.48\linewidth]{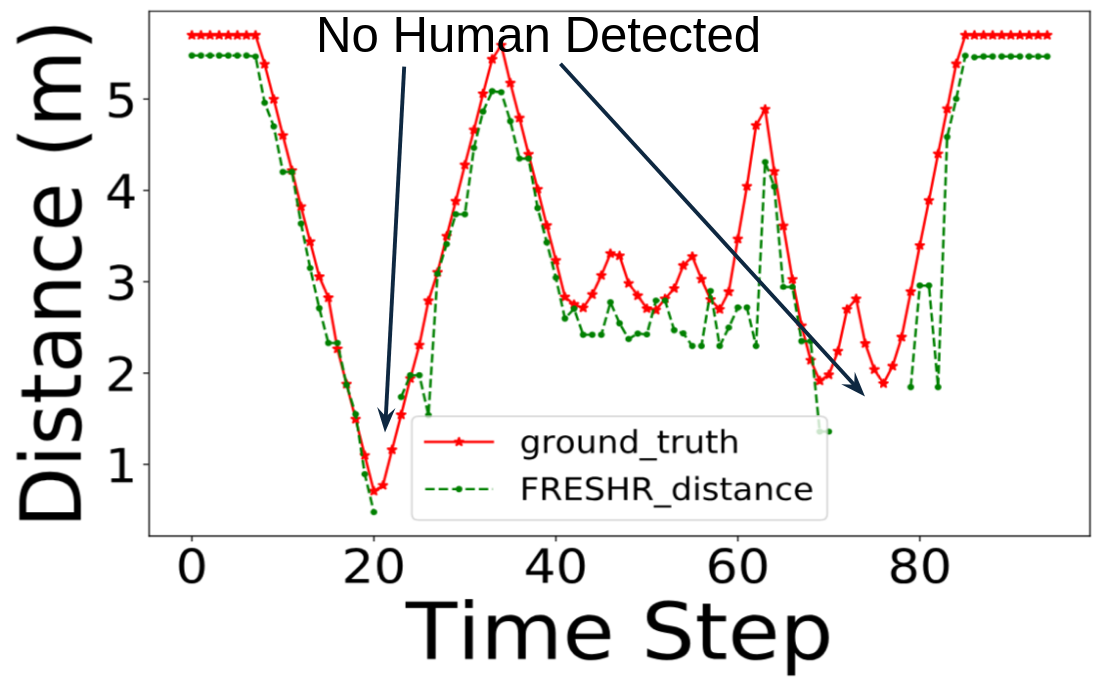} &
    \includegraphics[width=0.48\linewidth]{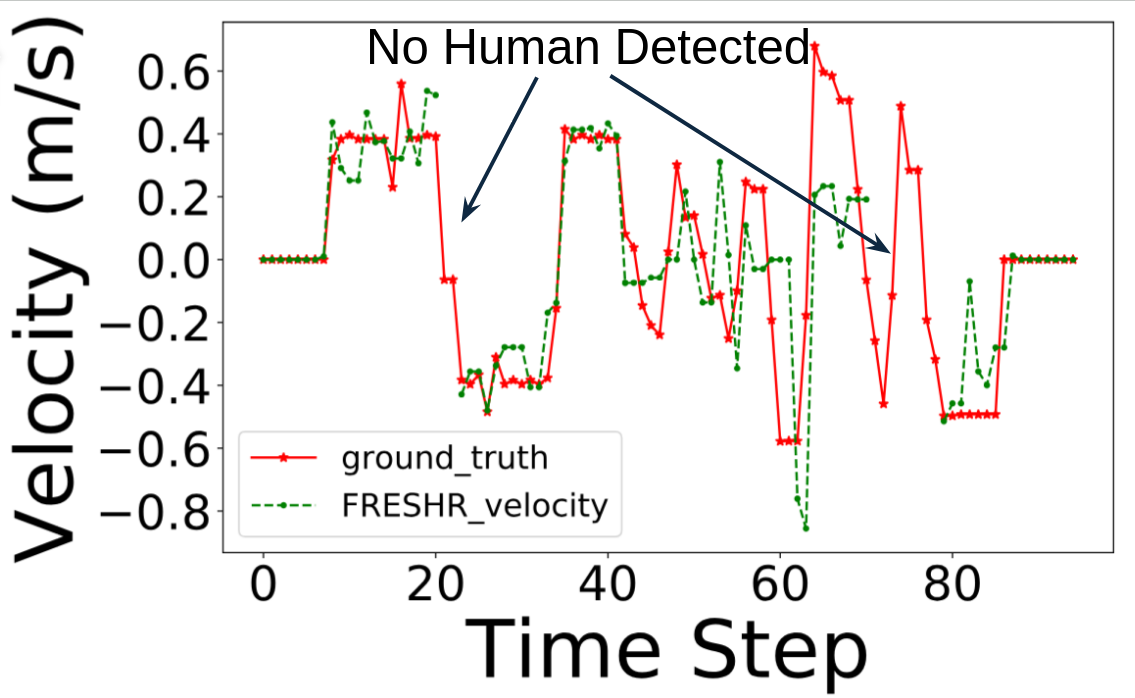} \\
    \multicolumn{2}{c}{\textbf{\small Observer's viewpoint}} \vspace{-1mm}\\
    \includesvg[width=0.48\linewidth]{case2_f.svg} & 
    \includesvg[width=0.48\linewidth]{case2_vel_f.svg}\\  
    \end{tabular}
    \centerline{\small ($b$)}
    \vspace{-2mm}
    \caption{\small ($a$) Human moves along three trajectories in the shared space: straight toward the task robot (setting 1), perpendicular to the task robot (setting 2), and diagonal (setting 3). ($b$) FRESHR-based estimation of the distances (left) and relative velocities (Right) from two different viewpoints.}
    \label{fig:relative_dist_vel}
    \vspace{-4mm}
\end{figure}

\subsection{Validating FRESHR Measurements}
\label{sec:result}

As the accuracy of GSI is contingent on correctly measuring the human's distance and relative velocity, we begin by assessing the error in obtaining these using typical sensors. We compare the FRESHR-provided distance and relative velocity with ground truth for a single human as he follows the trajectories shown in Fig.~\ref{fig:relative_dist_vel}($a$) in ROS Gazebo 11. The perceived distance and velocity measurements are obtained using the RealSense camera D435i model provided by Intel and YOLO v7. Figure~\ref{fig:relative_dist_vel} shows that we can follow the change in distance in real time (at a 30Hz rate) with reasonable accuracy. The mean absolute error in distance measurements are 11.3\% and 5.07\% from the task robot and the observer's viewpoints, respectively. Similarly, we observed the following error in the velocity measurements of the task robot at 16.6\% compared to the observer robot's at 9.61\%. The velocities measured by the task robot show some error (e.g., around $t=65$), and this is, in part, because the sensors and detection model do not perceive the human as stationary although the human's velocity is zero. This effect is diminished for the observer. The errors arise primarily due to the noisy readings of the camera's depth sensor. Importantly, we expect GSI from an observer's viewpoint to be much more accurate than from the task robot's viewpoint because an observer usually maintains a clear view of both the human and the task robot in its range. On the other hand, a moving human may not stay in the camera view of a navigating or stationary robot.

\begin{figure}[!t]
    \centering
    \begin{subfigure}[t]{0.49\linewidth}
    \raisebox{-\height}{\includegraphics[width=\linewidth]{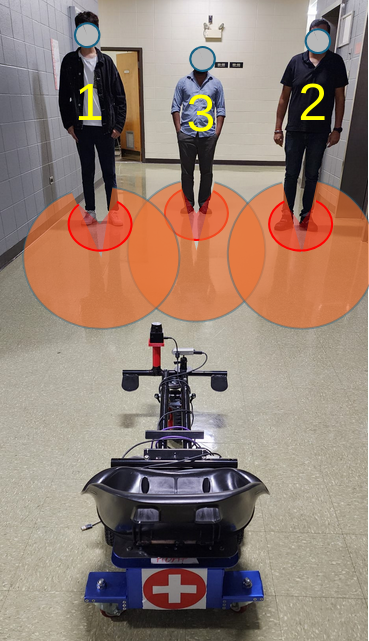} }    
    \centerline{\small ($a$)}
    \end{subfigure}
    \begin{subfigure}[t]{0.49\linewidth}
    \raisebox{-\height}{\includegraphics[width=\linewidth]{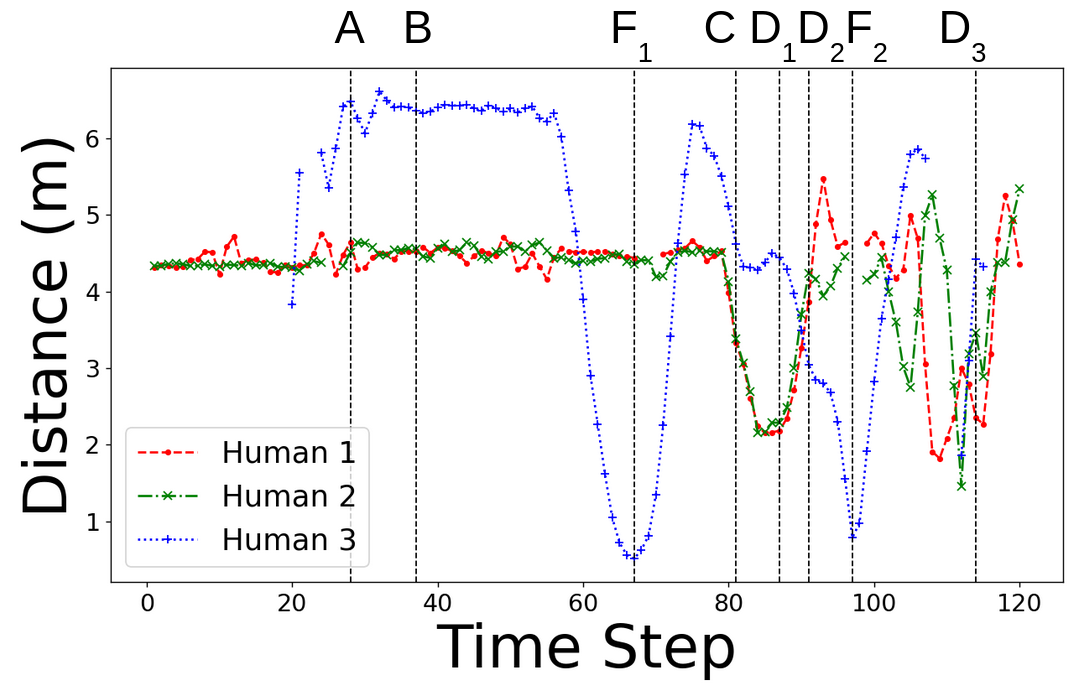} }
    \raisebox{-\height}{\includegraphics[width=\linewidth]{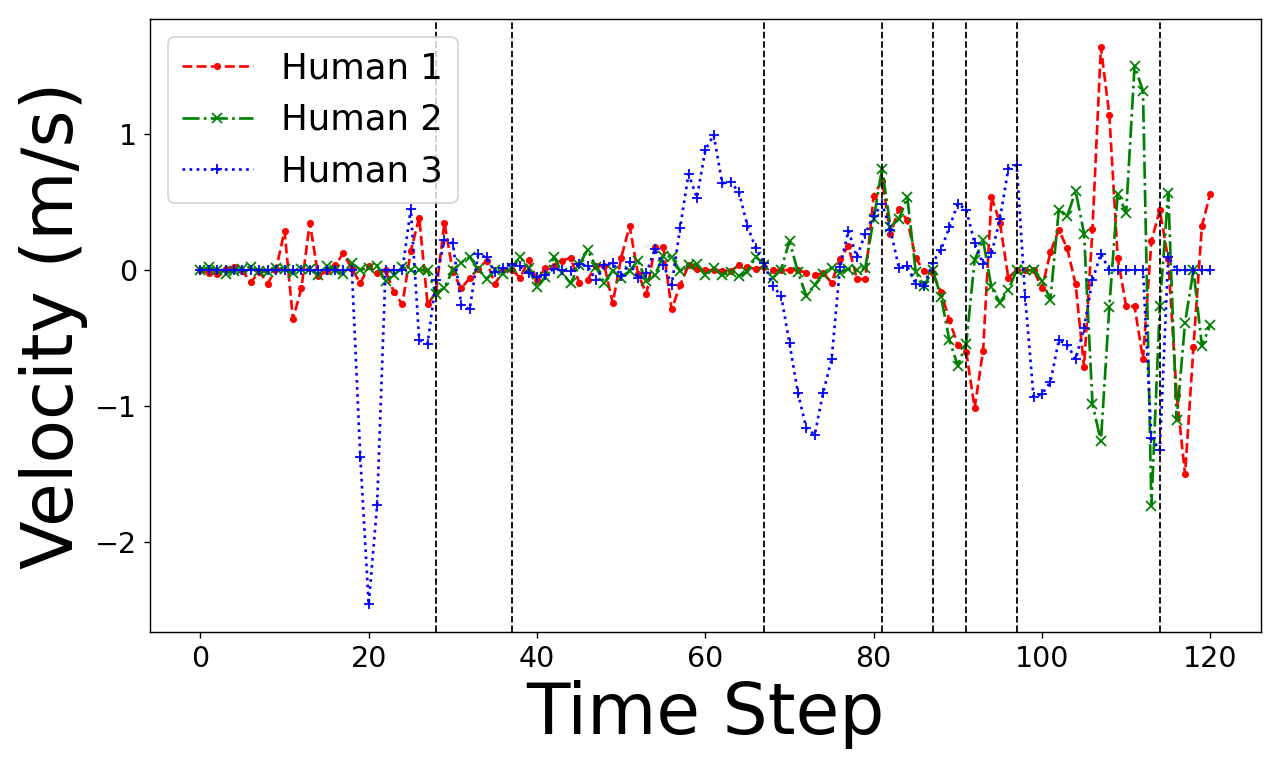} }
    \raisebox{-\height}{\includegraphics[width=\linewidth]{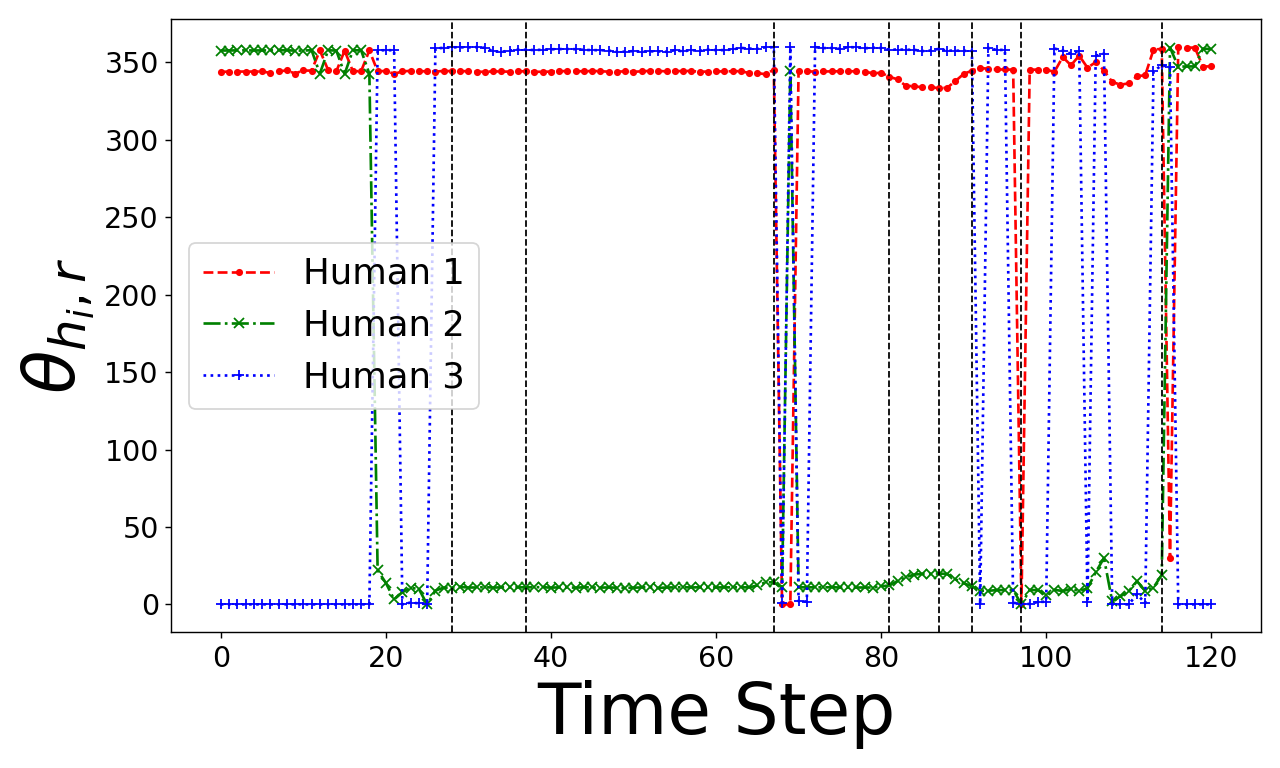} }
    \centerline{\small ($b$)}
    \end{subfigure}
\caption{\small ($a$) Setting for the physical robot experiment involving FRESHR on an Ubiquity Magni which is sharing space with three moving humans. ($b$) Validation of distances ${\bm{d}_{h,r}}$, relative velocities ${\bm{v}_{h,r}}$, and bearings ${\bm{\theta}_{h,r}}$ of the three humans.}
    \label{fig:mh_icra}
    \vspace{-5mm}
\end{figure}

\subsection{Validating Multi-Human Safety Assessment with GSI}
\label{sec:experiments}

We conducted physical robot experiments with a Ubiquity Magni platform customized for use in a medical evacuation application \cite{jordan2024analyzing} and equipped with an Intel RealSense D435i mounted in front, as we show in Fig.~\ref{fig:mh_icra}($a$). We created a multi-human scenario with three humans in the robot's view simultaneously. While the robot was stationary, the humans followed various trajectories at regular walking speeds to simulate a pedestrian walkway. This included: 
1) Human 3 alone walks toward the robot and then moves away. Other humans stay put;
2) Humans 1 and 2 walk toward the robot while Human 3 stays put;
3) Human 3 walks toward the robot while Humans 1 and 2 walk away;
and 4) Random movement of all humans.
The attached video contains a detailed demonstration of this experiment and the results. 

We show the distances, velocities, and bearings of the three humans engaged in these behaviors in Fig.~\ref{fig:mh_icra}, as measured by FRESHR. Scenarios A -- F presented in Table~\ref{tab:scenarios_appropriateness} manifest in these behaviors. These are marked in the three plots of Fig.~\ref{fig:mh_icra}($b$). We note that the measures correctly track as the humans move in the shared space leading to the scenarios.    

\begin{figure}[t]
\centering
        \includegraphics[width=0.99\linewidth]{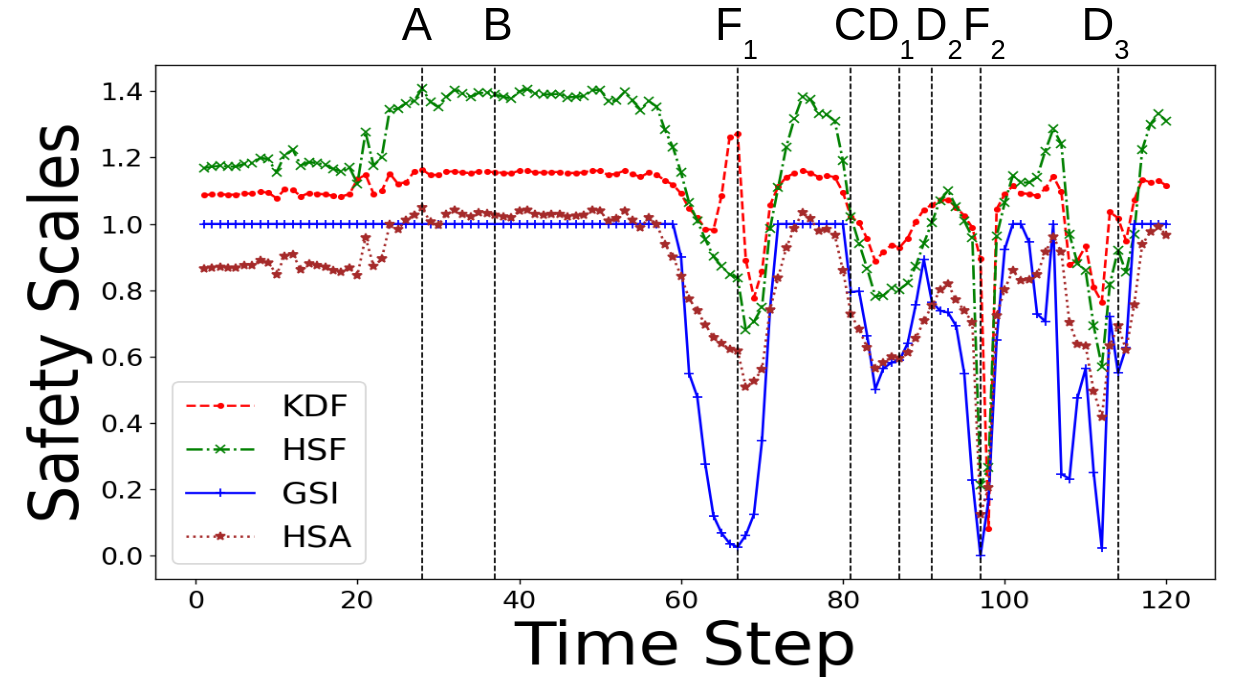} 
\caption{\small Output of safety scales applicable to multiple humans in our physical robot experiments with three humans.}
\label{fig:mh_comparison}
\vspace{-6mm}
\end{figure}

We compare the GSI safety scale (with $\rho=1$ and $\tau = 0.01$ as it is a service robot) with the existing kineostatic danger field (KDF) \cite{lacevic2013safety}, human safety field (HSF) \cite{palmieri2024control}, and human safety assessment (HSA) \cite{lippi2018safety} as these allow extensions to multiple humans, but were not explicitly tested in their respective studies. 
KDF, HSF, and HSA use averaging to aggregate the safety of multiple humans. We inverted the KDF values (as it is a danger scale, similar to DI) and scaled HSF by, $\frac{1}{Dmax}$ as it relies solely on the distance factor.


{\em Observe from Fig.~\ref{fig:mh_comparison} that KDF and HSF report safety values that are much higher than GSI.} This is because of the averaging utilized by these scales that generally lift safety when several humans are safe (but not all). For example, as Human 3 approaches the robot but Humans 1 and 2 stay put, which corresponds to scenario F$_1$, KDF, HSF, and HSA reduce but not as much as GSI. The latter's overall safety assessment emphasizes approaching humans over others. 

Another stark distinction between the four safety assessments is in scenarios D$_1$ and F$_2$, when Humans 1 and 2 are walking away while Human 3 is approaching (D$_1$) and when just Human 3 remains in the robot's viewable range and the robot is nearing the human's intimate space (F$_2$). While all assessments drop, the impact of Humans 1 and 2 walking back is much more on KDF and HSF while GSI remains sensitive to the approaching human. On the other hand, the presence of a robot in close proximity to the human in F$_2$ causes all scales to report low safety values. 

Based on this detailed analysis, we note that GSI behaves differently from extant scales KDF, HSF, and HSA in contexts involving multiple humans. Indeed, KDF and HSF appear to consistently overestimate the overall safety of the situation whereas GSI offers more specificity in assessing the safety of the multi-human situation.

\subsection{Evaluating the Utility of FRESHR and GSI}
\label{Sec: Socialnavigation}

\begin{figure}[!t]
    \centering
    \begin{subfigure}[t]{0.45\linewidth}
    \raisebox{-\height}{\includegraphics[width=\linewidth]{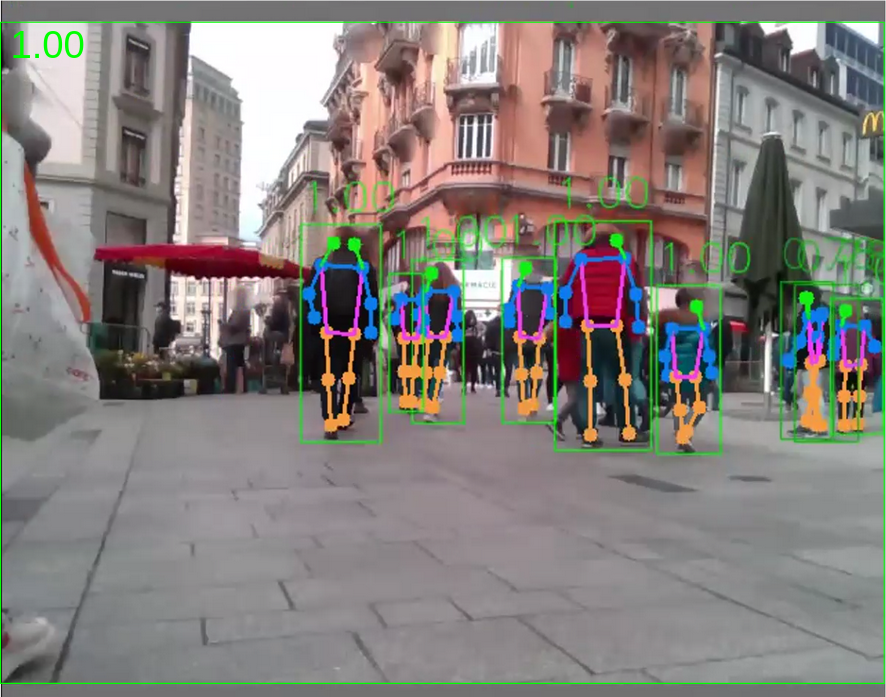} }
    \raisebox{-\height}{\includegraphics[width=\linewidth]{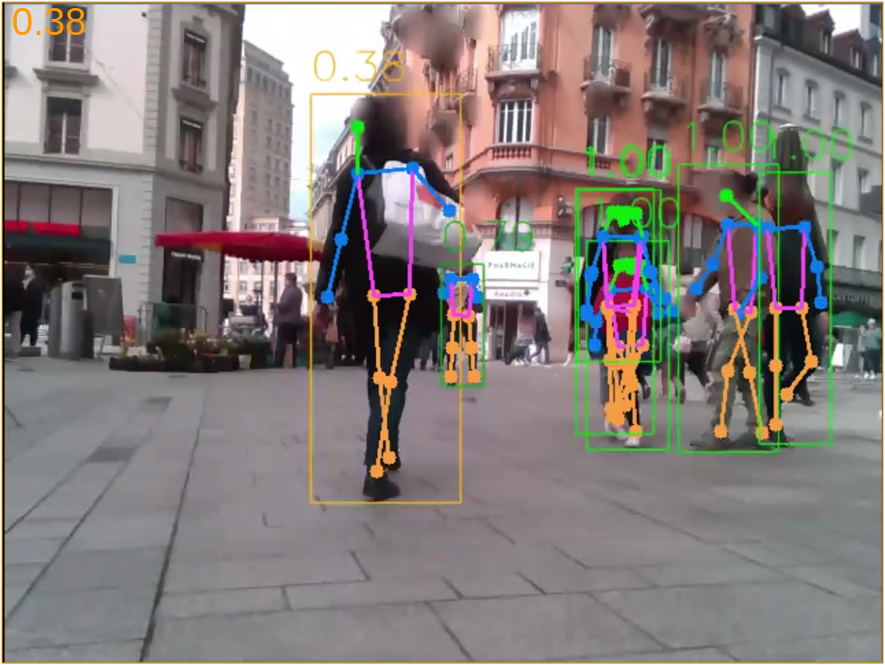} }
    \raisebox{-\height}{\includegraphics[width=\linewidth]{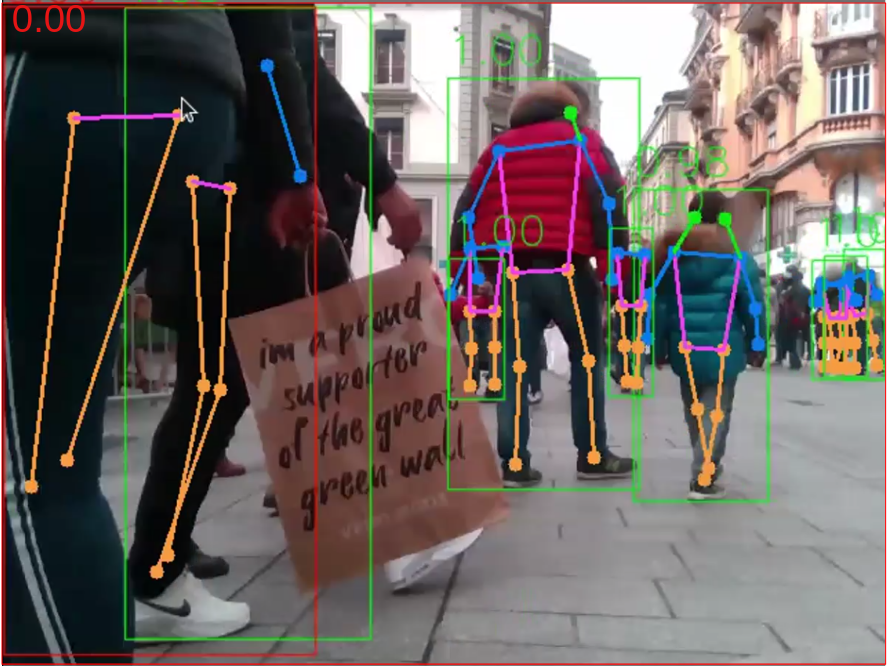} }
    \centerline{\small ($a$)}
    \end{subfigure}
     \begin{subfigure}[t]{0.53\linewidth}
    \raisebox{-\height}{\includegraphics[width=\linewidth]{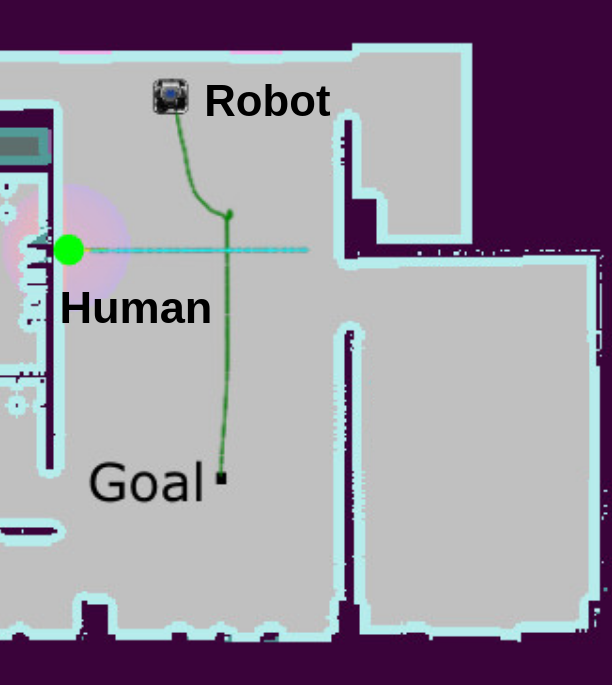} } 
    \raisebox{-\height}{\includegraphics[width=\linewidth]{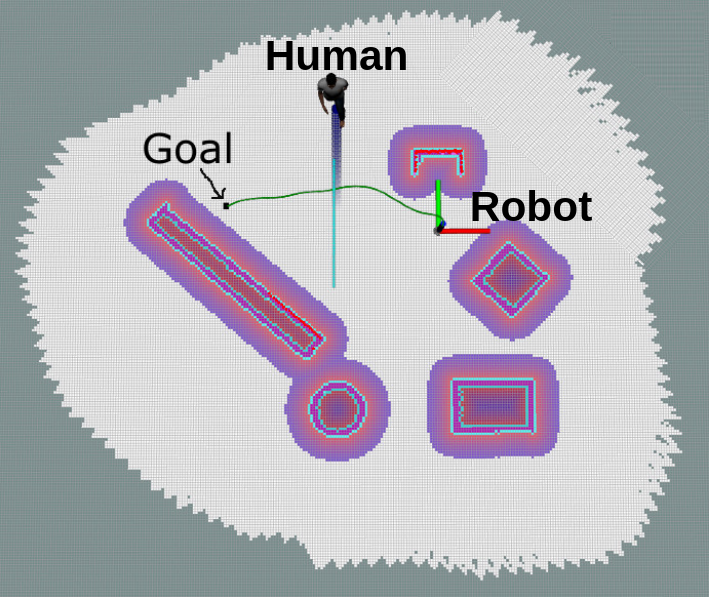} }  
    \centerline{\small ($b$)}
    \end{subfigure}
\caption{\small ($a$) Instances from EPFL Crowdbot dataset video of Shared Control, illustrating when our GSI indicates safe, between, and unsafe conditions. ($b$) Map and example trajectories of the human and robot in the human-aware HATEB \cite{hateb} (top) and regular A* \cite{kollmitz2015time} (below) environments in ROS Stage simulations. }
    \label{fig:mh_icra_pictures}
    \vspace{-6mm}
\end{figure}

We applied our framework to EPFL's Crowdbot
datasets \cite{paez2022pedestrian} and demonstrate the efficacy of using GSI to compare different motion planners for an assistive robot. We compare three different motion planners on this dataset: autonomous (MDS), manual, and shared control (SC) (see Fig.~\ref{fig:mh_icra_pictures}(a)). Shared control obtained the highest GSI of 0.82 and is the safest -- it balances human and robotic guidance allowing for intervention when needed. In comparison, autonomous with a GSI of 0.77 prioritizes efficiency and may bring the robot closer to humans, whereas manual control, with the lowest GSI of 0.53, often lacks safety boundaries.

Next, we compare a {\em human-aware} motion planner (HATEB \cite{hateb}) and a regular motion planner (A* that treats a human as an obstacle) from a safety perspective in settings shown in Fig.~\ref{fig:mh_icra_pictures}(b). Interestingly, HATEB showed a lower human safety with an average GSI of 0.73, compared to the average of 0.82 for A*. However, this is not surprising because human-aware motion planners are able to navigate closely around humans without invading their intimate space, whereas regular planners, perceiving humans as obstacles, refrain from crossing the social space around them.

\section{Conclusion}
\label{Sec: conclusion}

We presented a new model to assess human safety for mobile robots operating in a multi-human environment, with guidelines for configuring the safety scale. An RGB-D camera-based safety evaluation framework, FRESHR, uses the model to perform real-time safety assessments and allows multiple endpoint usage. Realistic simulations and physical robot experiments confirmed the validity of the model and its utility compared with other extant safety scales. The contributions in this work will help advance safety-aware algorithms and motion planners in human-rich mobile robot applications.

\section*{Acknowledgements} 
Research was sponsored by the DEVCOM Analysis Center and was accomplished under Cooperative Agreement Number W911NF-22-2-0001. The views and conclusions contained in this document are those of the authors. They should not be interpreted as representing the official policies, expressed or implied, of the DEVCOM Analysis Center or the U.S. Government. The U.S. Government is authorized to reproduce and distribute reprints for Government purposes, notwithstanding any copyright notation herein.
\bibliographystyle{IEEEtran}
\bibliography{newreferences}

\begin{thebibliography}{10}
\providecommand{\url}[1]{#1}
\csname url@samestyle\endcsname
\providecommand{\newblock}{\relax}
\providecommand{\bibinfo}[2]{#2}
\providecommand{\BIBentrySTDinterwordspacing}{\spaceskip=0pt\relax}
\providecommand{\BIBentryALTinterwordstretchfactor}{4}
\providecommand{\BIBentryALTinterwordspacing}{\spaceskip=\fontdimen2\font plus
\BIBentryALTinterwordstretchfactor\fontdimen3\font minus \fontdimen4\font\relax}
\providecommand{\BIBforeignlanguage}[2]{{%
\expandafter\ifx\csname l@#1\endcsname\relax
\typeout{** WARNING: IEEEtran.bst: No hyphenation pattern has been}%
\typeout{** loaded for the language `#1'. Using the pattern for}%
\typeout{** the default language instead.}%
\else
\language=\csname l@#1\endcsname
\fi
#2}}
\providecommand{\BIBdecl}{\relax}
\BIBdecl

\bibitem{lasota2017survey}
P.~A. Lasota, T.~Fong, J.~A. Shah \emph{et~al.}, ``A survey of methods for safe human-robot interaction,'' \emph{Foundations and Trends{\textregistered} in Robotics}, vol.~5, no.~4, pp. 261--349, 2017.

\bibitem{veloso2018increasingly}
M.~M. Veloso, ``The increasingly fascinating opportunity for human-robot-ai interaction: The cobot mobile service robots,'' pp. 1--2, 2018.

\bibitem{khalid2017safety}
A.~Khalid, P.~Kirisci, Z.~Ghrairi, J.~Pannek, and K.-D. Thoben, ``Safety requirements in collaborative human--robot cyber-physical system,'' in \emph{Dynamics in logistics}.\hskip 1em plus 0.5em minus 0.4em\relax Springer, 2017, pp. 41--51.

\bibitem{valori2021validating}
M.~Valori, A.~Scibilia, I.~Fassi, J.~Saenz, R.~Behrens, S.~Herbster, C.~Bidard, E.~Lucet, A.~Magisson, L.~Schaake \emph{et~al.}, ``Validating safety in human--robot collaboration: Standards and new perspectives,'' \emph{Robotics}, vol.~10, no.~2, p.~65, 2021.

\bibitem{matheson2019human}
E.~Matheson, R.~Minto, E.~G. Zampieri, M.~Faccio, and G.~Rosati, ``Human--robot collaboration in manufacturing applications: A review,'' \emph{Robotics}, vol.~8, no.~4, p. 100, 2019.

\bibitem{harper2010towards}
C.~Harper and G.~Virk, ``Towards the development of international safety standards for human robot interaction,'' \emph{International Journal of Social Robotics}, vol.~2, no.~3, pp. 229--234, 2010.

\bibitem{palmieri2024control}
J.~Palmieri, P.~Di~Lillo, M.~Lippi, S.~Chiaverini, and A.~Marino, ``A control architecture for safe trajectory generation in human--robot collaborative settings,'' \emph{IEEE Transactions on Automation Science and Engineering}, 2024.

\bibitem{nertinger2023influence}
S.~Nertinger, R.~J. Kirschner, S.~Abdolshah, A.~Naceri, and S.~Haddadin, ``Influence of robot motion and human factors on users' perceived safety in hri,'' in \emph{2023 IEEE International Conference on Advanced Robotics and Its Social Impacts (ARSO)}.\hskip 1em plus 0.5em minus 0.4em\relax IEEE, 2023, pp. 46--52.

\bibitem{ferraguti2020control}
F.~Ferraguti, M.~Bertuletti, C.~T. Landi, M.~Bonf{\`e}, C.~Fantuzzi, and C.~Secchi, ``A control barrier function approach for maximizing performance while fulfilling to iso/ts 15066 regulations,'' \emph{IEEE Robotics and Automation Letters}, vol.~5, no.~4, pp. 5921--5928, 2020.

\bibitem{kulic2007pre}
D.~Kuli{\'c} and E.~Croft, ``Pre-collision safety strategies for human-robot interaction,'' \emph{Autonomous Robots}, vol.~22, no.~2, pp. 149--164, 2007.

\bibitem{lippi2018safety}
M.~Lippi and A.~Marino, ``Safety in human-multi robot collaborative scenarios: a trajectory scaling approach,'' \emph{IFAC-PapersOnLine}, vol.~51, no.~22, pp. 190--196, 2018.

\bibitem{hall1966hidden}
E.~T. Hall, ``The hidden dimension,'' \emph{Garden City}, 1966.

\bibitem{kamide2014direct}
H.~Kamide, Y.~Mae, T.~Takubo, K.~Ohara, and T.~Arai, ``Direct comparison of psychological evaluation between virtual and real humanoids: Personal space and subjective impressions,'' \emph{International Journal of Human-Computer Studies}, vol.~72, no.~5, pp. 451--459, 2014.

\bibitem{rossi2017user}
S.~Rossi, M.~Staffa, L.~Bove, R.~Capasso, and G.~Ercolano, ``User’s personality and activity influence on hri comfortable distances,'' in \emph{Social Robotics: 9th International Conference, ICSR 2017, Tsukuba, Japan, November 22-24, 2017, Proceedings 9}.\hskip 1em plus 0.5em minus 0.4em\relax Springer, 2017, pp. 167--177.

\bibitem{yasumoto2011personal}
M.~Yasumoto, H.~Kamide, Y.~Mae, K.~Kawabe, S.~Sigemi, M.~Hirose, and T.~Arai, ``Personal space of humans in relation with humanoid robots depending on the presentation method,'' in \emph{2011 IEEE/SICE International Symposium on System Integration (SII)}.\hskip 1em plus 0.5em minus 0.4em\relax IEEE, 2011, pp. 797--801.

\bibitem{brandl2016human}
C.~Brandl, A.~Mertens, and C.~M. Schlick, ``Human-robot interaction in assisted personal services: factors influencing distances that humans will accept between themselves and an approaching service robot,'' \emph{Human Factors and Ergonomics in Manufacturing \& Service Industries}, vol.~26, no.~6, pp. 713--727, 2016.

\bibitem{leichtmann2020much}
B.~Leichtmann and V.~Nitsch, ``How much distance do humans keep toward robots? literature review, meta-analysis, and theoretical considerations on personal space in human-robot interaction,'' \emph{Journal of environmental Psychology}, vol.~68, p. 101386, 2020.

\bibitem{steinfeld2006common}
A.~Steinfeld, T.~Fong, D.~Kaber, M.~Lewis, J.~Scholtz, A.~Schultz, and M.~Goodrich, ``Common metrics for human-robot interaction,'' in \emph{Proceedings of the 1st ACM SIGCHI/SIGART conference on Human-robot interaction}, 2006, pp. 33--40.

\bibitem{truong2017toward}
X.-T. Truong and T.~D. Ngo, ``Toward socially aware robot navigation in dynamic and crowded environments: A proactive social motion model,'' \emph{IEEE Transactions on Automation Science and Engineering}, vol.~14, no.~4, pp. 1743--1760, 2017.

\bibitem{halme2018review}
R.-J. Halme, M.~Lanz, J.~K{\"a}m{\"a}r{\"a}inen, R.~Pieters, J.~Latokartano, and A.~Hietanen, ``Review of vision-based safety systems for human-robot collaboration,'' \emph{Procedia CIRP}, vol.~72, pp. 111--116, 2018.

\bibitem{rodrigues2022modeling}
I.~R. Rodrigues, G.~Barbosa, A.~Oliveira~Filho, C.~Cani, M.~Dantas, D.~H. Sadok, J.~Kelner, R.~S. Souza, M.~V. Marquezini, and S.~Lins, ``Modeling and assessing an intelligent system for safety in human-robot collaboration using deep and machine learning techniques,'' \emph{Multimedia Tools and Applications}, vol.~81, no.~2, pp. 2213--2239, 2022.

\bibitem{maria2022vision}
L.~Mar{\'\i}a Amaya-Mej{\'\i}a, N.~Duque-Su{\'a}rez, D.~Jaramillo-Ram{\'\i}rez, and C.~Martinez, ``Vision-based safety system for barrierless human-robot collaboration,'' \emph{arXiv e-prints}, pp. arXiv--2208, 2022.

\bibitem{svarny2019safe}
P.~Svarny, M.~Tesar, J.~K. Behrens, and M.~Hoffmann, ``Safe physical hri: Toward a unified treatment of speed and separation monitoring together with power and force limiting,'' in \emph{2019 IEEE/RSJ International Conference on Intelligent Robots and Systems (IROS)}.\hskip 1em plus 0.5em minus 0.4em\relax IEEE, 2019, pp. 7580--7587.

\bibitem{tashtoush2021human}
T.~Tashtoush, L.~Garcia, G.~Landa, F.~Amor, A.~N. Laborde, D.~Oliva, and F.~Safar, ``Human-robot interaction and collaboration (hri-c) utilizing top-view rgb-d camera system,'' \emph{International Journal of Advanced Computer Science and Applications}, vol.~12, no.~1, 2021.

\bibitem{secil2022minimum}
S.~Secil and M.~Ozkan, ``Minimum distance calculation using skeletal tracking for safe human-robot interaction,'' \emph{Robotics and Computer-Integrated Manufacturing}, vol.~73, p. 102253, 2022.

\bibitem{lacevic2013safety}
B.~Lacevic, P.~Rocco, and A.~M. Zanchettin, ``Safety assessment and control of robotic manipulators using danger field,'' \emph{IEEE Transactions on Robotics}, vol.~29, no.~5, pp. 1257--1270, 2013.

\bibitem{lasota2014toward}
P.~A. Lasota, G.~F. Rossano, and J.~A. Shah, ``Toward safe close-proximity human-robot interaction with standard industrial robots,'' in \emph{2014 IEEE International Conference on Automation Science and Engineering (CASE)}.\hskip 1em plus 0.5em minus 0.4em\relax IEEE, 2014, pp. 339--344.

\bibitem{edelmann2023interaction}
A.~Edelmann, S.~St{\"u}mper, and T.~Petzoldt, ``The interaction between perceived safety and perceived usefulness in automated parking as a result of safety distance,'' \emph{Applied ergonomics}, vol. 108, 2023.

\bibitem{salvini2022safety}
P.~Salvini, D.~Paez-Granados, and A.~Billard, ``Safety concerns emerging from robots navigating in crowded pedestrian areas,'' \emph{International Journal of Social Robotics}, vol.~14, no.~2, pp. 441--462, 2022.

\bibitem{ferrer2013robot}
G.~Ferrer, A.~Garrell, and A.~Sanfeliu, ``Robot companion: A social-force based approach with human awareness-navigation in crowded environments,'' in \emph{2013 IEEE/RSJ International Conference on Intelligent Robots and Systems}.\hskip 1em plus 0.5em minus 0.4em\relax IEEE, 2013, pp. 1688--1694.

\bibitem{kulic2006real}
D.~Kuli{\'c} and E.~A. Croft, ``Real-time safety for human--robot interaction,'' \emph{Robotics and Autonomous Systems}, vol.~54, no.~1, pp. 1--12, 2006.

\bibitem{wang2022yolov7}
C.-Y. Wang, A.~Bochkovskiy, and H.-Y.~M. Liao, ``Yolov7: Trainable bag-of-freebies sets new state-of-the-art for real-time object detectors,'' \emph{arXiv preprint arXiv:2207.02696}, 2022.

\bibitem{jordan2024analyzing}
T.~Jordan, P.~Pandey, P.~Doshi, R.~Parasuraman, and A.~Goodie, ``Analyzing human perceptions of a medevac robot in a simulated evacuation scenario,'' \emph{arXiv preprint arXiv:2410.19072}, 2024.

\bibitem{hateb}
P.~Teja~Singamaneni, A.~Favier, and R.~Alami, ``Human-aware navigation planner for diverse human-robot interaction contexts,'' in \emph{2021 IEEE/RSJ International Conference on Intelligent Robots and Systems (IROS)}, 2021, pp. 5817--5824.

\bibitem{kollmitz2015time}
M.~Kollmitz, K.~Hsiao, J.~Gaa, and W.~Burgard, ``Time dependent planning on a layered social cost map for human-aware robot navigation,'' in \emph{2015 European Conference on Mobile Robots (ECMR)}.\hskip 1em plus 0.5em minus 0.4em\relax IEEE, 2015, pp. 1--6.

\bibitem{paez2022pedestrian}
D.~Paez-Granados, Y.~He, D.~Gonon, D.~Jia, B.~Leibe, K.~Suzuki, and A.~Billard, ``Pedestrian-robot interactions on autonomous crowd navigation: Reactive control methods and evaluation metrics,'' in \emph{2022 IEEE/RSJ International Conference on Intelligent Robots and Systems (IROS)}.\hskip 1em plus 0.5em minus 0.4em\relax IEEE, 2022, pp. 149--156.

\end{thebibliography}

\end{document}